\documentclass[fleqn,10pt]{wlscirep}
\usepackage[utf8]{inputenc}
\usepackage[T1]{fontenc}
\usepackage{hyperref}       
\usepackage{url}            
\usepackage{booktabs}       
\usepackage{amsfonts}       
\usepackage{nicefrac}       
\usepackage{microtype}      
\usepackage{xcolor}         
\usepackage{booktabs}
\usepackage{diagbox}
\usepackage{enumitem}
\usepackage{threeparttable}
\usepackage{xspace}
\usepackage{adjustbox}
\usepackage{amsmath}
\usepackage{wrapfig}
\usepackage{lineno}
\usepackage{cleveref}
\usepackage{marvosym}
\usepackage{multirow}
\usepackage{longtable}
\usepackage{makecell}  
\usepackage[most]{tcolorbox}
\usepackage{etoolbox} 
\usepackage{setspace}
\usepackage{fontawesome5}
\usepackage{array}    
\usepackage{colortbl} 

\makeatletter
\pretocmd{\@bibitem}
  {\ifboolexpr{ test {\ifdefstring{\@biblabel}{added}} }{\color{red}}{}}
  {}{}
\makeatother

\crefname{figure}{Figure}{Figures}
\crefname{section}{Section}{Sections}
\crefname{table}{Table}{Tables}
\crefname{equation}{Equation}{Equations}
\crefname{appendix}{Appendix}{Appendixes}

\newcommand{\benchmark}{\textsc{BioVista}\xspace}
\newcommand{\framework}{\textsc{BioMiner}\xspace}
\newcommand{\MLLM}{\textsc{BioMiner-Instruct}\xspace}
\newcommand{\OCSR}{\textsc{MolGlyph}\xspace}

\definecolor{replyblue}{RGB}{0,0,0}
\definecolor{replybluesr}{RGB}{0, 0, 0} 

\newcommand{\yan}[1]{{\color{replyblue}{#1}}}
\newcommand{\yansr}[1]{{\color{replybluesr}{#1}}} 

\title{\framework: A Multi-modal System for Automated Mining of Protein-Ligand Bioactivity Data from Literature}

\author[1*]{Jiaxian Yan}
\author[2*]{Jintao Zhu}
\author[1]{Yuhang Yang}
\author[1,\Letter]{Qi Liu}
\author[1]{Kai Zhang}
\author[3]{Zaixi Zhang}
\author[1]{Xukai Liu}
\author[1]{Boyan Zhang}
\author[4]{Kaiyuan Gao}
\author[5]{Jinchuan Xiao}
\author[1]{Enhong Chen}

\affil[1]{State Key Laboratory of Cognitive Intelligence, University of Science and Technology of China}
\affil[2]{Center for Quantitative Biology, Academy for Advanced Interdisciplinary Studies, Peking University}
\affil[3]{Princeton University}
\affil[4]{Huazhong University of Science and
Technology}
\affil[5]{Infinite Intelligence Pharma}
\affil[*]{Jiaxian Yan and Jintao Zhu contribute equally to this work.}
\affil[\Letter]{Qi Liu is the corresponding author. qiliuql@ustc.edu.cn.}

\newtcolorbox{prompt}[1]{
    breakable,         
    enhanced,
    attach boxed title to top left={xshift=4mm, yshift=-2mm},
    colback=white,
    colframe=replyblue!70,
    coltitle=white,
    boxrule=0.5pt,
    arc=2pt,
    left=6pt, right=6pt, top=8pt, bottom=6pt,
    fontupper=\scriptsize\sffamily,
    title={\bfseries\small #1}, 
    boxed title style={
        sharp corners, 
        rounded corners=northwest, 
        colback=replyblue!70, 
        boxrule=0pt,
        frame hidden,
    },
    drop shadow=replyblue!30,
}

\definecolor{sopblue}{RGB}{0, 85, 128} 
\definecolor{sopbg}{RGB}{245, 250, 255} 

\newlist{steps}{enumerate}{1}
\setlist[steps]{
    label=\textbf{Step \arabic*.} , 
    leftmargin=*,
    labelsep=0.5em,
    itemsep=0pt, 
    topsep=0pt,
    partopsep=0pt,
    font=\sffamily\small
}

\newtcolorbox{sop}[1]{
    enhanced,
    breakable,
    attach boxed title to top left={xshift=4mm, yshift=-2mm},
    colback=replyblue!5,          
    colframe=replyblue!75!black,  
    coltitle=white,
    boxrule=0.5pt,
    arc=2pt,
    left=5pt, right=5pt, top=6pt, bottom=5pt,
    fontupper=\small\linespread{0.95}\selectfont,
    title={\bfseries\small SOP: #1},
    boxed title style={
        sharp corners,
        rounded corners=northwest,
        colback=replyblue!75!black, 
        boxrule=0pt,
        frame hidden,
    },
    drop shadow=black!30,
}

\begin{abstract}
Protein-ligand bioactivity data published in the literature are essential for drug discovery, yet manual curation struggles to keep pace with rapidly growing literature. \yan{Automated bioactivity extraction remains challenging because it requires not only interpreting biochemical semantics distributed across text, tables, and figures, but also reconstructing chemically exact ligand structures (e.g., Markush structures).
To address this bottleneck, we introduce \framework, a multi-modal extraction framework that explicitly separates bioactivity semantic interpretation from ligand structure construction. Within \framework, bioactivity semantics are inferred through direct reasoning, while chemical structures are resolved via a chemical-structure-grounded visual semantic reasoning paradigm, in which multi-modal large language models operate on chemically grounded visual representations to infer inter-structure relationships, and exact molecular construction is delegated to domain chemistry tools. For rigorous evaluation and method development, we further establish \benchmark,} a comprehensive benchmark comprising 16,457 bioactivity entries curated from 500 publications. \framework validates its extraction ability and provides a quantitative baseline, achieving an F1 score of 0.32 for bioactivity triplets. \framework's practical utility is demonstrated via three applications: (1) extracting 82,262 data from 11,683 papers to build a pre-training database that improves downstream  models performance by 3.9\%; (2) enabling a human-in-the-loop workflow that doubles the number of high-quality NLRP3 bioactivity data, helping 38.6\% improvement over 28 QSAR models and identification of 16 hit candidates with novel scaffolds; and (3) \yansr{accelerating protein-ligand complex bioactivity annotation, achieving a 5.59-fold speed increase and 5.75\% accuracy improvement over manual workflows in PoseBusters dataset.}
\framework and \benchmark provide a scalable extraction methodology and a rigorous benchmark, paving the way to unlock bioactivity data that previously required extensive human effort.
\yansr{All data and code are available at \href{https://github.com/jiaxianyan/BioMiner}{GitHub}.}
\end{abstract}
\begin{document}

\flushbottom

\maketitle

\thispagestyle{empty}

\vspace{-0.1cm}
\section*{Main}

Protein-ligand bioactivity data represent a cornerstone of modern drug discovery~\cite{Gaulton2011ChEMBLAL, Zhang2025StructureBasedDD}, \yan{underpinning structure-activity relationships (SAR) analysis~\cite{Theisen2024LeveragingMD, Lai2024InterformerAI, Shah2025DeepDTAGenAM, Lu2025DTIAMAU, Ye2021AUD, Koh2023PSICHICPG, Mastropietro2023LearningCO, Zhang2024EfficientGO, Feng2024ABF}, quantitative structure-activity relationship (QSAR) modeling, and AI-driven virtual screening~\cite{Gentile2022ArtificialIV, Cao2024GenericPI}.
Despite the availability of large public resources such as ChEMBL~\cite{zdrazil2024chembl}, BindingDB~\cite{liu2025bindingdb}, and PDBbind~\cite{Liu2015PDBwideCO}, the continual expansion of these databases still relies predominantly on manual expert curation.
This reliance has emerged as a fundamental scalability bottleneck, increasingly unable to keep pace with the exponential growth of the scientific literature.
}

Automated data extraction tools have been developed in various scientific domains, yet a critical gap exists in specialized tools tailored for the extraction of protein-ligand bioactivity data. 
Early tools, constrained by the limitations of prior natural language processing (NLP) and computer vision (CV) techniques, primarily focused on tasks of extracting fundamental chemical information, such as named entity recognition (NER)~\cite{Lan2024GeneratingMO,song2021deep, Dagdelen2024StructuredIE} and optical chemical structure recognition (OCSR)~\cite{Morin2024PatCIDAO}. 
Recently, the development of large language models (LLMs) has advanced some more promising text-mining tools. For example, GPT-based tools have been proposed for extracting metal-organic framework synthesis conditions and enzyme-substrate interactions from literature~\cite{Zheng2023ChatGPTCA, Smith2024FuncFetchAL, Kang2024ChatMOFAA}.

\yan{However, these methods, while demonstrating progress, are not yet equipped to handle the extraction of protein-ligand bioactivity data (Table S1), which requires jointly resolving biochemical semantics and chemically exact structural representations. 
This intrinsic coupling fundamentally distinguishes bioactivity extraction from conventional scientific information extraction and gives rise to three core challenges.
\textbf{First}, bioactivity data is inherently multi-modal, distributed across different modalities, including text, tables, figures, and crucial chemical structures.
This multi-modal challenge requires robust cross-modal reasoning rather than unimodal extraction.}
\textbf{Second}, the accurate recognition and conversion of chemical structures, particularly the widely used Markush structures (which represent groups of related chemical compounds)~\cite{Simmons2003MarkushSS}, remains a significant obstacle. While existing OCSR methods have the ability to recognize some Markush scaffold or R-group substitute structures, the crucial step of enumerating the specific, individual compounds they represent—a necessity for precise bioactivity data extraction—has been largely unexplored.
\yan{\textbf{Third}, besides these intrinsic data complexities, the field} lacks standardized, large-scale benchmarks. This critical gap severely hinders the rigorous evaluation, comparative analysis, and systematic advancement of automated methodologies, impeding the development and validation of robust, generalizable solutions capable of reliably unlocking the wealth of bioactivity data embedded within the scientific literature.

\begin{figure}[!htbp]
\centering
\includegraphics[width=1.0\linewidth]{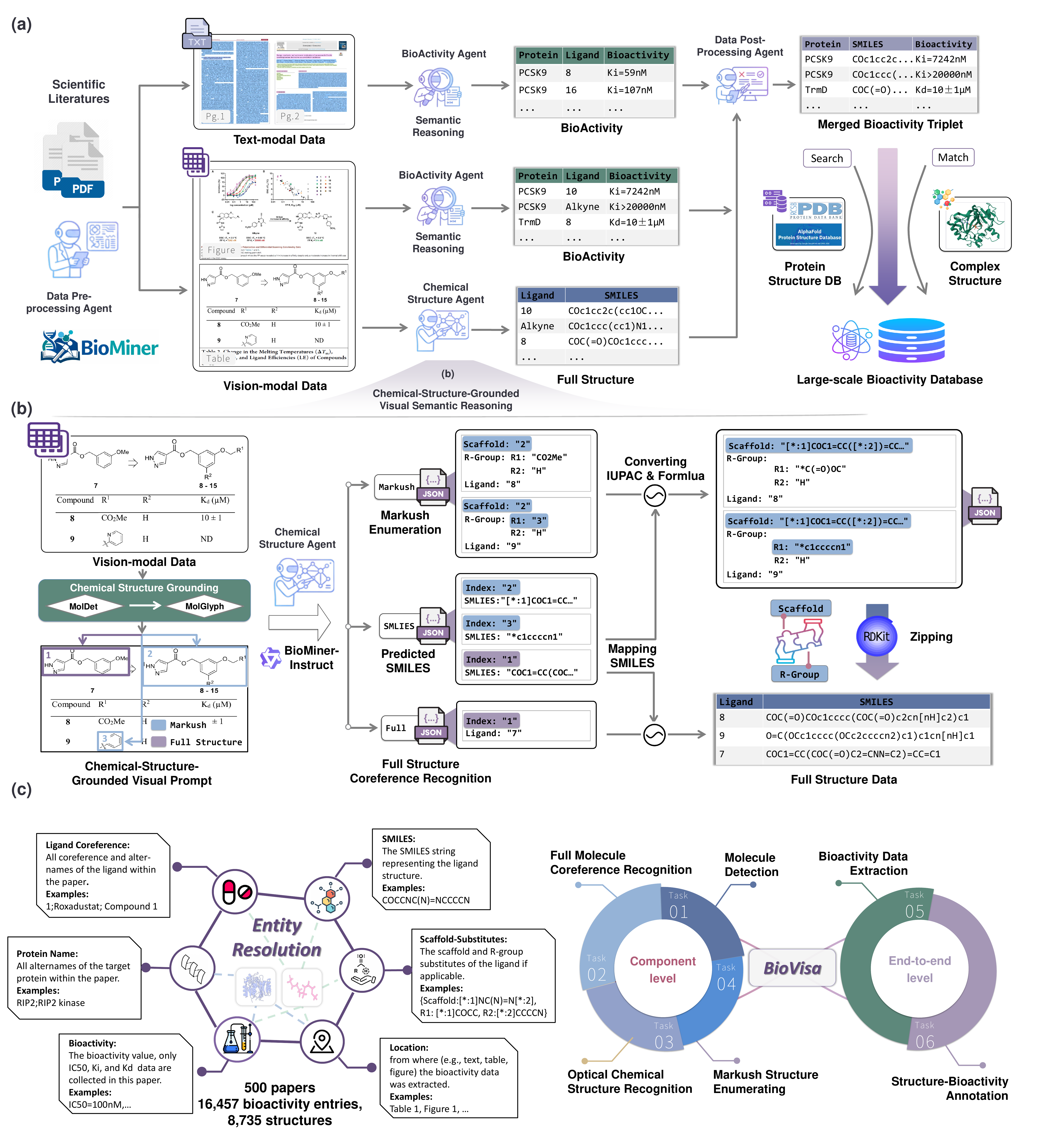}
\caption{\yan{Overview of protein-ligand bioactivity extraction framework \framework and benchmark \benchmark.
(a) The whole protein-ligand bioactivity extraction framework \framework.
(b) The chemical structure extraction agent. In this agent, explicit full structures and Markush structures are both processed. For clarification, explicit full structures are plotted in \yansr{purple} boxes, and Markush scaffolds and R-group substituents are plotted in blue boxes. 
(c) The new benchmark \benchmark, containing 16,457 bioactivity data and 8,735 structures as ground-truth labels from 500 publications. Based on these data, six evaluation tasks are designed for comprehensive evaluation, ranging from \yansr{component-level tasks to end-to-end tasks}. }
}
\label{fig:overall_framewprk}
\end{figure}

To overcome these challenges, we propose \textbf{\framework}, \yan{a multi-modal agentic system designed to automatically extract protein-ligand bioactivity data from literature.
Rather than relying on one-shot end-to-end extraction, \framework decomposes the extraction task into document parsing, chemical structure extraction, bioactivity measurement extraction, and cross-modal integration, with each subtask handled by specialized agents.
As illustrated in Figure~\ref{fig:overall_framewprk}\yansr{(a)}, the two core subtasks (bioactivity measurement extraction and chemical structure extraction) are explicitly separated and addressed using fundamentally different strategies, where bioactivity measurements are directly extracted through semantic reasoning, whereas chemical structures are resolved via a chemical-structure-grounded visual semantic reasoning (CSG-VSR) mechanism.
This design is motivated by their intrinsic difference: bioactivity measurement extraction is dominated by biochemical semantic reasoning, while chemical structure extraction additionally requires the exact construction of chemically valid symbolic representations.
We therefore decouple chemical semantic reasoning from strict symbolic construction and integrate \yansr{multi-modal LLM (MLLM)} reasoning with \yansr{domain-specific models (DSMs)} and chemistry tools for chemical symbolic processing. Such a strategy enables the resolution of complex chemical representations, particularly enumerating complex Markush structures into specific full molecular structures—a pivotal step previously unaddressed at scale for automated bioactivity extraction (Figure~\ref{fig:overall_framewprk}(b)).
}

To provide standardized resources for evaluation and support future research, we \yan{further} introduce benchmark \textbf{\benchmark}, to our knowledge, the largest benchmark dedicated to protein-ligand bioactivity extraction. 
Meticulously curated by domain experts, \benchmark comprises 16,457 bioactivity entries and 8,735 unique chemical structures extracted from 500 recent publications indexed in PDBbind v2020~\cite{Liu2015PDBwideCO}. 
The data originates from diverse modalities within the source papers: text (15.8\%), figures (11.6\%), and tables (72.5\%), with a substantial fraction of chemical structures (48.7\%) derived from challenging Markush representations.
\benchmark supports six distinct evaluation tasks, ranging from end-to-end bioactivity extraction to component-level Markush enumeration, allowing for comprehensive evaluation~(Figure~\ref{fig:overall_framewprk}(c)).
\yan{To enable rigorous and unbiased evaluation, \benchmark is constructed with a strictly held-out test set and a separate validation set used exclusively for model development and ablation, preventing any form of test-time tuning or leakage.
When evaluated on \benchmark, \framework achieves an F1 score of 0.32 for extracting complete bioactivity triplets.
}

We further perform three applications to showcase \framework's practical utility and broad applicability.
\textbf{First}, \yan{we use \framework to construct a bioactivity pre-training database containing 82,262 data points extracted from 11,683 papers within three days. Models pre-trained on this \framework-generated database demonstrate improved performance (3.9\% and 3.4\% improvement in RMSE metric) on two independent test sets (PDBbind v2016 core set~\cite{Su2018ComparativeAO} and CSAR-HiQ~\cite{Dunbar2013CSARDS}) compared to models trained only on existing curated data (PDBbind v2016 refined set, 3,767 data points).}
\textbf{Second}, we implement a human-in-the-loop (HITL) bioactivity extraction workflow, where human experts collect data by reviewing the extraction result of \framework rather than \textit{de novo} identification and transcription, to correct errors of fully automated extraction and ensure data quality. With the HITL workflow, 1,592 bioactivity data of NLRP3~\cite{Swanson2019TheNI}, a high-priority target for anti-inflammatory therapies, are collected from 85 papers in 26 hours, doubling the NLRP3 data available in ChEMBL. This expanded dataset yields improved QSAR models (38.6\% improvement in EF1\% metric), based on which we screen ChemDiv~\cite{chemdiv} and Enamine~\cite{enamine} virtual compound libraries and identify 16 hit candidates with novel scaffolds.
\textbf{Third}, besides bioactivity extraction, we utilize \framework to label complex structures with reported bioactivity data, which is important in structure-based drug design for establishing datasets like PDBbind.
\yansr{Evaluated on 242 complexes from the PoseBusters database~\cite{Buttenschoen2023PoseBustersAD}, \framework-assisted workflow outperformed fully manual annotation across 2 crossover analyses (4 annotators), improving the average accuracy from 90.5\% to 96.25\% and reducing the average annotation time from 195.8\,s to 35.0\,s (5.59-fold faster).}
These experiments, taken together, demonstrate \framework's efficiency, applicability, and potential to accelerate drug discovery through both fully automated and HITL workflows. In summary, \framework and \benchmark provide a new system and evaluation standard for automated bioactivity data extraction. These contributions offer a pathway to unlock vast amounts of previously inaccessible bioactivity data, accelerating data-driven drug discovery and establishing a foundation for future progress in automated scientific literature mining.

\vspace{-0.1cm}
\section*{Results}
To elucidate the performance and utility of our proposed framework, \framework, we first briefly outline its architecture, with a focus on the core chemical structure extraction agent that underpins its ability to handle complex structure data (including Markush structures). Next, we introduce \benchmark, to the best of our knowledge, the largest benchmark designed to rigorously evaluate bioactivity extraction performance.
We provide a detailed analysis of \framework's extraction performance on diverse tasks of \benchmark. Finally, we demonstrate \framework's practical value through three real-world applications, highlighting its versatility and impact in benefiting bioactivity data extraction and drug design.

\vspace{-0.1cm}
\subsection*{Framework \framework}
\yan{\framework, depicted in Figure~\ref{fig:overall_framewprk}(a), is a multi-modal agentic system specially designed for protein-ligand bioactivity extraction.}

\subsubsection*{Overview of \framework}
\yan{Rather than introducing \framework as a collection of agents or models, we first clarify the design principle that governs the entire system.
The central challenge in automated protein–ligand bioactivity extraction lies in two fundamentally different requirements:
(1) semantic reasoning over heterogeneous, multi-modal bioactivity evidence, and
(2) the exact construction of chemically valid ligand structures, particularly Markush structures.
End-to-end extraction approaches entangle these requirements, rendering the task structurally brittle, particularly in the presence of complex chemical representations.

Instead of performing a one-shot end-to-end prediction, \framework explicitly \yansr{decouples} bioactivity semantic interpretation from ligand structure construction.
Specifically, \framework~\yansr{decomposes} the task into four stages---document parsing, bioactivity measurement interpretation, chemical structure resolution, and cross-modal integration---each corresponding to a distinct source of uncertainty and handled by specialized components.
\yansr{The document is first parsed with MinerU~\cite{Wang2024MinerUAO}, after which bioactivity measurement extraction and chemical structure resolution proceed in parallel.
Both branches are built around \MLLM, a domain-specialized MLLM fine-tuned from Qwen3-VL-32B~\cite{Bai2025Qwen3VLTR}.
Bioactivity measurements are extracted separately from text, tables, and figures using a post-fusion strategy.
We adopt this strategy in \framework and optimize \MLLM for this setting based on our empirical result (Figure~S12(a)), where the general MLLM Gemini-2.0-flash achieves better extraction performance under post-fusion.
In the other branch,} ligand structures are resolved through a \yansr{CSG-VSR} strategy, described in detail below, which anchors MLLM reasoning to detected chemical depictions and enforces chemical validity through \yansr{DSMs} and \yansr{chemistry} tools.
Finally, a post-processing agent integrates semantic bioactivity \yansr{measurements} with resolved ligand structures to produce complete bioactivity triplets.}

\subsubsection*{\textcolor{replyblue}{Chemical-Structure-Grounded Visual Semantic Reasoning}}
\yan{Chemical structure resolution constitutes the primary structural bottleneck in automated bioactivity extraction, particularly for Markush representations that encode combinatorial chemical spaces rather than explicit molecules.
To address this challenge, \framework introduces a CSG-VSR mechanism, which anchors high-level MLLM reasoning to chemically grounded visual and symbolic representations.

CSG-VSR operates in three stages. 
First, \yansr{DSMs} detect and parse 2D chemical depictions from figures and tables. MolDetv2~\cite{huggingface_repo_moldetv2} identifies molecular structure regions, followed by \yansr{OCSR} using \OCSR (details in Methods section) to generate SMILES representations. This stage provides a chemically grounded visual substrate that constrains subsequent reasoning.
Second, an MLLM performs visual semantic reasoning over augmented images containing indexed chemical depictions. For explicit full structures, the model resolves coreference between textual mentions and detected depictions. For Markush structures, the model identifies the Markush scaffold and semantically enumerates associated R-group definitions, which may be expressed visually, textually, or symbolically. Crucially, the MLLM is responsible only for relational and semantic reasoning (e.g., scaffold–substituent associations), not for enforcing chemical validity.
Third, chemical symbolic construction is carried out deterministically using domain tools to systematically zip the recognized Markush scaffold SMILES with the enumerated R-group substituent SMILES, generating the final list of specific, full chemical structures represented by the Markush definition. 
Notably, R-group substituents are often described textually (e.g., IUPAC names, abbreviations, chemical formulas) rather than visually. Before the RDKit zipping step, these 1D R-group substituents are converted into SMILES, employing OPSIN~\cite{Lowe2011ChemicalNT} for IUPAC names and a Gemini-assisted, manually curated mapping table for abbreviations and chemical formulas.

By decoupling semantic reasoning from chemical symbolic construction and grounding both stages in domain-specific representations, CSG-VSR enables scalable and reliable resolution of complex Markush structures without task-specific model training.
This capability is essential for automated bioactivity extraction at scale and underpins the performance gains observed in subsequent evaluations.}

\begin{figure}[!tbp]
\centering
\includegraphics[width=1.0\linewidth]{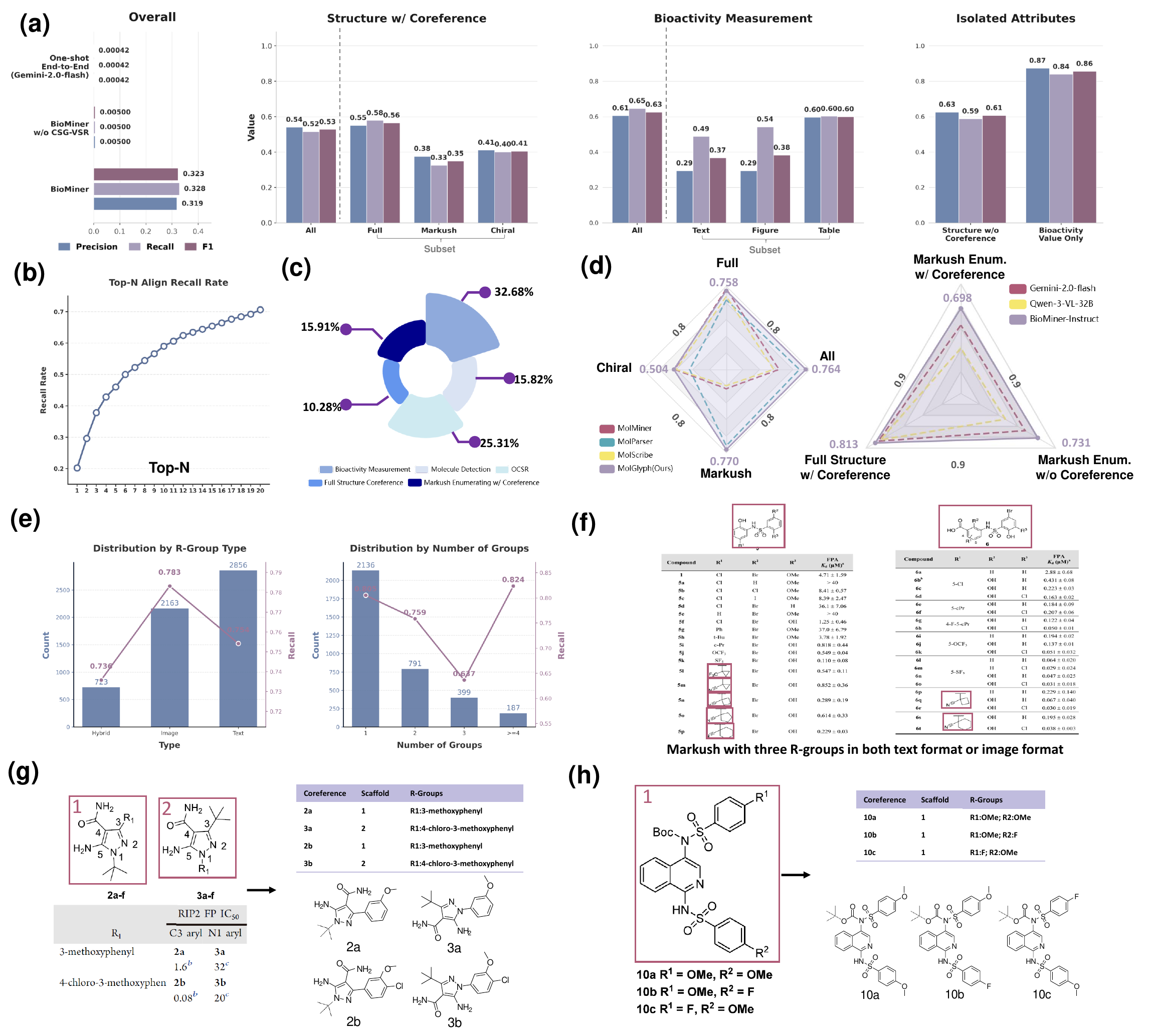}
\caption{\yansr{Benchmarking extraction performance of \framework on \benchmark.
(a) Performance of bioactivity triplet and individual attributes extraction. For overall extraction performance, one-shot end-to-end extraction baseline and \framework w/o CSG-VSR ablation study are included additionally.  
(b) Performance of the structure-bioactivity annotation task.
(c) Detailed error source analysis of bioactivity triplet extraction. 
(d) Component-level performance of OCSR models (left) and MLLM models (right).
(e) Quantitative analysis of Markush enumeration recall under varying R-group modalities (left) and substitution complexity (right).
(f) Failed examples of Markush enumeration with three R-groups.
(g,h) Two examples of Markush structures successfully processed by \framework.}
}
\label{fig:extracting_performance}
\end{figure}

\subsection*{Benchmark \benchmark and Performance Evaluation}
To enable rigorous evaluation and facilitate the development of future extraction methods, we introduce \benchmark, a comprehensive and challenging benchmark for protein-ligand bioactivity data extraction. 
To our knowledge, this is the largest benchmark dedicated to the bioactivity extraction task.
In this subsection, we first present the construction process of \benchmark, and then detail \framework's performance.

\subsubsection*{Construction of \benchmark}
\benchmark is derived from 500 recent publications referenced in PDBbind v2020~\cite{Liu2015PDBwideCO}, ensuring its relevance to real-world scenarios. 
\yan{While sourced from PDBbind, our benchmark represents a diverse and challenging distribution of scientific literature. Analysis shows that these 500 publications cover 102 distinct journals, exhibiting a "long-tail" distribution unlike standard databases (e.g., ChEMBL), which are heavily concentrated in a few top-tier journals (Figure S2). This ensures the model is evaluated on a wide variety of layout styles and textual reporting habits.
}
Unlike PDBbind data curation, \benchmark includes all bioactivity data reported within these publications, not just the reported bioactivity to the PDB structure.
Domain experts manually \yan{(details about experts in Table S3)} curate all bioactivity data points, annotating attributes beyond the core triplet (protein, ligand SMILES, bioactivity value) to enable fine-grained analysis (Figure~\ref{fig:overall_framewprk}(c)).
These attributes (e.g., location, ligand coreference, scaffold-substitutes) include information such as the source modality to analyze the ability to process different modal data. 
Notably, all alternative names (alternames) for proteins and ligands are carefully collected, ensuring completeness and preventing ambiguity or overlap.

\yan{To ensure high data quality and completeness, we employed a model-assisted verification strategy aimed at identifying potential omissions (false negatives) in the initial human annotations. Specifically, the \framework was utilized to scan the curated publications, and its extractions were cross-referenced with human annotations. Any discrepancies or newly identified data points were rigorously reviewed by domain experts. Only data points with explicit location verified by humans were added to the ground truth, ensuring the dataset remains objective and free from model-induced hallucinations. 
After annotation, we conduct\yansr{ed} an Inter-Annotator Agreement study~\cite{lombard2002content} on a random 10\% subset \yansr{(}50 papers\yansr{)} to analyze the quality of \benchmark. Such analysis yield\yansr{ed} an F1 \yansr{score} of 0.899, indicating high consistency and reliability.}
The finalized \benchmark dataset comprises 16,457 bioactivity data points sourced from text (15.8\%), figures (11.6\%), and tables (72.5\%) (Figure~S1), along with 8,735 chemical structures (of which 48.7\% are derived from Markush structures). 
To strictly prevent "tuning on the test set" and ensure unbiased evaluation, \benchmark \yansr{ papers were} randomly partitioned into a Validation Set (50 papers, 10\%) and a held-out Test Set (450 papers, 90\%). 
\yansr{We employed the 10\%/90\% split ratio to prioritize a more stable estimate of final model performance on unseen papers.}
All prompt engineering, hyperparameter tuning, and rule refinement described in this study were conducted exclusively based on the Validation Set. The Test Set remained unseen \yansr{for \framework} and was used solely for the performance evaluation.

\benchmark defines two end-to-end tasks and four component-level tasks to thoroughly assess the extraction performance of \framework. The two end-to-end tasks evaluate overall extraction capability, specifically: (1) extracting all bioactivity data reported in a publication, and (2) annotating PDB structures with bioactivity information presented in associated papers. These tasks directly reflect the practical utility of bioactivity extraction methods. Additionally, four specialized component-level tasks—molecule detection, OCSR, full structure coreference resolution, and Markush enumeration—provide deeper insights into the performance of critical chemical structure extraction processes, facilitating method development and optimization. Detailed descriptions of these evaluation tasks can be found in \yansr{Supplementary} Note 1. \yansr{Together, \benchmark serves not only as a comprehensive benchmark for \framework, but also as a valuable foundation for} advancing bioactivity extraction systems.

\begin{table}[t]
 \centering
 \caption{Performance of \framework with different MLLMs on \benchmark. This table focuses on presenting MLLMs-related results and comparing the performance of different MLLMs. \yansr{Note that recall is reported at multiple top ranks for the bioactivity-structure annotation task.} The best results are bolded, and the second best results are \underline{underlined.}}
 \label{tbl:overall_performance}
\begin{adjustbox}{width=1.0\textwidth}
\begin{threeparttable}
\begin{tabular}{
    >{\color{black}}c 
    >{\color{black}}c|
    >{\color{replyblue}}c
    >{\color{replyblue}}c 
    >{\color{replyblue}}c 
    >{\color{replyblue}}c 
    >{\color{replyblue}}c
    >{\color{replyblue}}c
    >{\color{replyblue}}c 
    >{\color{replyblue}}c 
}
\toprule
Tasks & Metric & \begin{tabular}[c]{@{}c@{}}\underline{Gemini-}\\ \underline{2.0-flash}\end{tabular} & GPT-4o-mini & GPT-4.1-mini & Grok-4-fast & \begin{tabular}[c]{@{}c@{}}Claude-\\ haiku-4-5\end{tabular} & \begin{tabular}[c]{@{}c@{}}Qwen3-\\ VL-32B\end{tabular} & \begin{tabular}[c]{@{}c@{}}Qwen3-VL-\\ 235B\end{tabular} & \begin{tabular}[c]{@{}c@{}}\textbf{\textsc{BioMiner-}}\\ \textbf{\textsc{Instruct}}\end{tabular} \\
\midrule
\multicolumn{5}{l}{\textit{Component-level evaluation for chemical structure extraction}} & & & & & \\
\midrule
\multirow{3}{*}{\begin{tabular}[c]{@{}c@{}}Full Structure\\ Coreference\end{tabular}} 
& Recall    & \underline{0.786} & 0.552 & 0.781 & 0.559 & 0.722 & 0.766 & 0.785 & \textbf{\textcolor{replyblue}{0.815}}\\
& Precision & \underline{0.783} & 0.550 & 0.777 & 0.558 & 0.719 & 0.762 & 0.782 & \textbf{\textcolor{replyblue}{0.812}}\\
& F1        & \underline{0.784} & 0.551 & 0.779 & 0.558 & 0.720 & 0.764 & 0.783 & \textbf{\textcolor{replyblue}{0.813}}\\

\midrule
\multirow{3}{*}{\begin{tabular}[c]{@{}c@{}}Markush Enum.\\ w/ Coreference\end{tabular}} 
& Recall    & 0.579 & 0.039 & 0.219 & 0.168 & 0.309 & 0.375 & \underline{0.611} & \textbf{\textcolor{replyblue}{0.709}}\\
& Precision & 0.545 & 0.041 & 0.220 & 0.162 & 0.289 & 0.369 & \underline{0.576} & \textbf{\textcolor{replyblue}{0.687}}\\
& F1        & 0.561 & 0.040 & 0.219 & 0.165 & 0.299 & 0.372 & \underline{0.593} & \textbf{\textcolor{replyblue}{0.698}}\\
\midrule
\multirow{3}{*}{\begin{tabular}[c]{@{}c@{}}Markush Enum.\\ w/o Coreference\end{tabular}} 
& Recall    & 0.628 & 0.050 & 0.255 & 0.223 & 0.380 & 0.422 & \underline{0.655} & \textbf{\textcolor{replyblue}{0.743}}\\
& Precision & 0.594 & 0.058 & 0.287 & 0.212 & 0.363 & 0.422 & \underline{0.618} & \textbf{\textcolor{replyblue}{0.719}}\\
& F1        & 0.611 & 0.054 & 0.270 & 0.217 & 0.371 & 0.422 & \underline{0.636} & \textbf{\textcolor{replyblue}{0.731}}\\
\midrule

\multicolumn{5}{l}{\textit{Different attributes evaluation for end-to-end extraction and annotation}} & & & & & \\
\midrule

\multirow{3}{*}{\begin{tabular}[c]{@{}c@{}}Structure\\ w/ Coreference\end{tabular}} 
& Recall    & \underline{0.482} & 0.310 & 0.413 & 0.413 & 0.269 & 0.406 & 0.462 & \textbf{\textcolor{replyblue}{0.516}}\\
& Precision & \underline{0.520} & 0.360 & 0.482 & 0.468 & 0.312 & 0.489 & 0.498 & \textbf{\textcolor{replyblue}{0.541}}\\
& F1        & \underline{0.500} & 0.333 & 0.445 & 0.439 & 0.289 & 0.444 & 0.479 & \textbf{\textcolor{replyblue}{0.528}}\\

\midrule
\multirow{3}{*}{\begin{tabular}[c]{@{}c@{}}Bioactivity\\ Measurement\end{tabular}} 
& Recall    & 0.583 & 0.251 & 0.363 & 0.513 & 0.400 & \underline{0.628} & 0.611 & \textbf{\textcolor{replyblue}{0.647}}\\
& Precision & \underline{0.522} & 0.236 & 0.422 & 0.368 & 0.255 & 0.501 & 0.423 & \textbf{\textcolor{replyblue}{0.606}}\\
& F1        & 0.551 & 0.243 & 0.390 & 0.429 & 0.311 & \underline{0.557} & 0.500 & \textbf{\textcolor{replyblue}{0.626}}\\

\midrule

\multirow{3}{*}{\begin{tabular}[c]{@{}c@{}}Bioactivity\\ Triplet\end{tabular}} 
& Recall    & \underline{0.284} & 0.089 & 0.145 & 0.225 & 0.112 & 0.253 & 0.271 & \textbf{\textcolor{replyblue}{0.328}}\\
& Precision & \underline{0.260} & 0.082 & 0.171 & 0.174 & 0.073 & 0.202 & 0.199 & \textbf{\textcolor{replyblue}{0.319}}\\
& F1        & \underline{0.272} & 0.085 & 0.157 & 0.196 & 0.088 & 0.224 & 0.230 & \textbf{\textcolor{replyblue}{0.323}}\\

\midrule
\multirow{3}{*}{\begin{tabular}[c]{@{}c@{}}Bioactivity-Structure\\ Annotation\end{tabular}} 
& Top-1     & 0.198 & \underline{0.218} & 0.192 & 0.198 & 0.214 & 0.198 & \textbf{0.238} & \textcolor{replyblue}{0.202} \\
& Top-3     & 0.362 & \textbf{0.458} & 0.400 & 0.372 & 0.400 & 0.376 & \underline{0.444} & \textcolor{replyblue}{0.378} \\
& Top-10    & 0.568 & \textbf{0.652} & 0.598 & 0.570 & \underline{0.626} & 0.586 & \textbf{0.652} & \textcolor{replyblue}{0.598} \\
\bottomrule
\end{tabular}
\end{threeparttable}
\end{adjustbox}
\end{table}

\subsubsection*{End-to-end Evaluation} 
\yansr{The} end-to-end bioactivity extraction performance is shown in Figure~\ref{fig:extracting_performance}(a). \framework achieves \yansr{a} precision \yansr{of} 0.319, \yansr{a} recall \yansr{of} 0.328, and \yansr{an} F1 \yansr{score of} 0.323 for bioactivity triplets. \yansr{In contrast, a one-shot end-to-end baseline that directly processes full-text and images attains an F1 score of 0.00042, highlighting the intrinsic difficulty of the task and the necessity of principled task decomposition.} 
\yansr{Granular analysis of individual attributes reveals even more robust performance compared to the integrated triplet.} Ligand coreference-SMILES extraction achieves an F1 \yansr{score of} 0.528, with performance varying by complexity  (Explicit Full F1 = 0.565 vs. Markush F1 = 0.349). 
Bioactivity measurement extraction (protein-ligand coreference-bioactivity value) achieves an F1 score of 0.626, with table-based extraction being the most effective (Table F1 = 0.600; Figure F1 = 0.382; Text F1 = 0.368). Isolated attribute extraction is \yansr{notably} proficient, \yansr{reaching F1 scores of 0.857 for bioactivity values and 0.606 for ligand structures}.
These \yansr{results} underscore the significant challenge of bioactivity extraction and indicate component strengths of \framework despite integration challenges.
\yansr{To quantify the specific contribution of CSG-VSR, an ablation study is conducted with the same \MLLM backbone. The removal of CSG-VSR leads to a sharp drop in the bioactivity triplet F1 from 0.323 to 0.011 (Figure~\ref{fig:extracting_performance}(a); Table~S8), confirming the effectiveness of the CSG-VSR mechanism.}

Performance on the structure-bioactivity annotation \yansr{task} is presented in Figure~\ref{fig:extracting_performance}(b). 
Given a complex structure and the associated paper, \framework extracts all bioactivity data from the paper, and ranks them based on ligand similarity between the extracted data and the given structure.
\yansr{Evaluation of the top-10 candidates reveals} a recall rate of 0.598 for the reported bioactivity  (Table~\ref{tbl:overall_performance}).
\yansr{The result suggests} that while the top-ranked match may not always be correct, the reported \yansr{bioactivity} of the given structure is \yansr{frequently ranked} among the top candidates, \yansr{thereby facilitating efficient automated annotation and HITL validation}.

\subsubsection*{Component-level Evaluation} 
\yansr{Chemical structure extraction, as the core module of \framework and a critical determinant of overall extraction fidelity, is rigorously assessed through four purpose-built component-level tasks.} As introduced above, \framework resolves ligand structures in three stages. \yansr{These tasks assess the first-stage molecule detection, OCSR, and the second-stage processes of full structure coreference resolution and Markush enumeration.}

\yansr{In the first stage, \framework utilizes MolDetv2 for molecule detection, yielding an mAP of 0.747, and }an AP$_{50}$ of 0.922 (Table~S6). 
\yansr{For OCSR, the \OCSR model employed herein surpasses existing methods, attaining superior overall accuracy (0.764) with a remarkable advantage in resolving Markush structures (0.770 accuracy; Table~S7).
Regarding second-stage tasks, \MLLM achieves F1 scores of 0.813 and 0.698 for full structure coreference resolution and Markush enumeration with coreference, respectively (Table~\ref{tbl:overall_performance}). The latter increases to 0.731 when coreference resolution is excluded, highlighting \MLLM's robustness in Markush enumeration.
Further benchmark study shows that \MLLM outperforms several general-purpose MLLMs, including Gemini, GPT~\cite{Hurst2024GPT4oSC}, Claude~\cite{anthropic_haiku45_systemcard}, Qwen~\cite{Bai2023QwenVLAV}, and Grok~\cite{Grok}, in both end-to-end extraction and component-level reasoning tasks (Table~\ref{tbl:overall_performance}). Collectively, \framework demonstrates leading performance in protein–ligand bioactivity extraction tasks owing to its efficient modular design.
}
\subsubsection*{Error Analysis and Cases Study} 
\yansr{We next analyze the component-wise error contribution within \framework's end-to-end bioactivity triplet extraction.}
Errors in bioactivity measurement extraction contribute the \yansr{predominant fraction} \yan{(32.68\%)}, followed by OCSR inaccuracies \yan{(25.31\%)}, and Markush enumeration failures \yan{(15.91\%)}. Molecule detection \yan{(15.82\%)} and explicit full structures coreference \yansr{resolution} \yan{(10.28\%)} contribute less to the overall error (Figure~\ref{fig:extracting_performance}(c)). This breakdown provides critical insights for directing future research toward optimizing the system components.

\yan{
We observe that, in addition to bioactivity extraction, errors arising from OCSR and Markush enumeration constitute the most substantial portions of the overall error budget. 
Accordingly, a more detailed analysis of these two components \yansr{is conducted}.
For OCSR, the results indicate that, beyond Markush structures, chirality recognition remains a major challenge. 
Even for our best-performing OCSR model, the accuracy on chiral structures is limited to 0.504, highlighting the intrinsic difficulty of precise stereochemical interpretation from heterogeneous visual inputs \yansr{(Figure~\ref{fig:extracting_performance}(d))}.
For Markush enumeration, we find that the recall is lowest when R-groups are jointly specified in both textual and graphical forms, suggesting that cross-modal alignment between textual R-group descriptions and visual depictions is error-prone \yansr{(Figure~\ref{fig:extracting_performance}(e))}.
Moreover, we observe a pronounced decline in recall when the number of R-groups increases to three, reflecting the combinatorial complexity introduced by multi-substituent enumeration \yansr{(Figure~\ref{fig:extracting_performance}(e, f))}.
These \yansr{results indicate that advancing stereochemical OCSR and cross-modal Markush enumeration will be pivotal for further improvement}.
}

Qualitative evidence of \framework's advanced capabilities is shown in \yansr{Figure~\ref{fig:extracting_performance}(g, h)}, which displays examples of complex Markush structures successfully processed by the system. This includes correct scaffold identification, R-group enumeration, 1D R-group processing (IUPAC \yansr{names and abbreviations}), and accurate generation of the final enumerated SMILES strings, indicating its ability to process challenging Markush structures.

In summary, the benchmarking results on \benchmark validate \framework's capacity for automated, multi-modal bioactivity data extraction. While quantifying the performance across various dimensions and pinpointing key challenges for future work, the results demonstrate certain bioactivity extraction ability and substantial efficiency advantages over manual curation. This positions \framework as a valuable tool for accelerating data acquisition and unlocking previously inaccessible data in the vast body of drug discovery literature.

\begin{figure}[!t]
\centering
\includegraphics[width=1.0\linewidth]{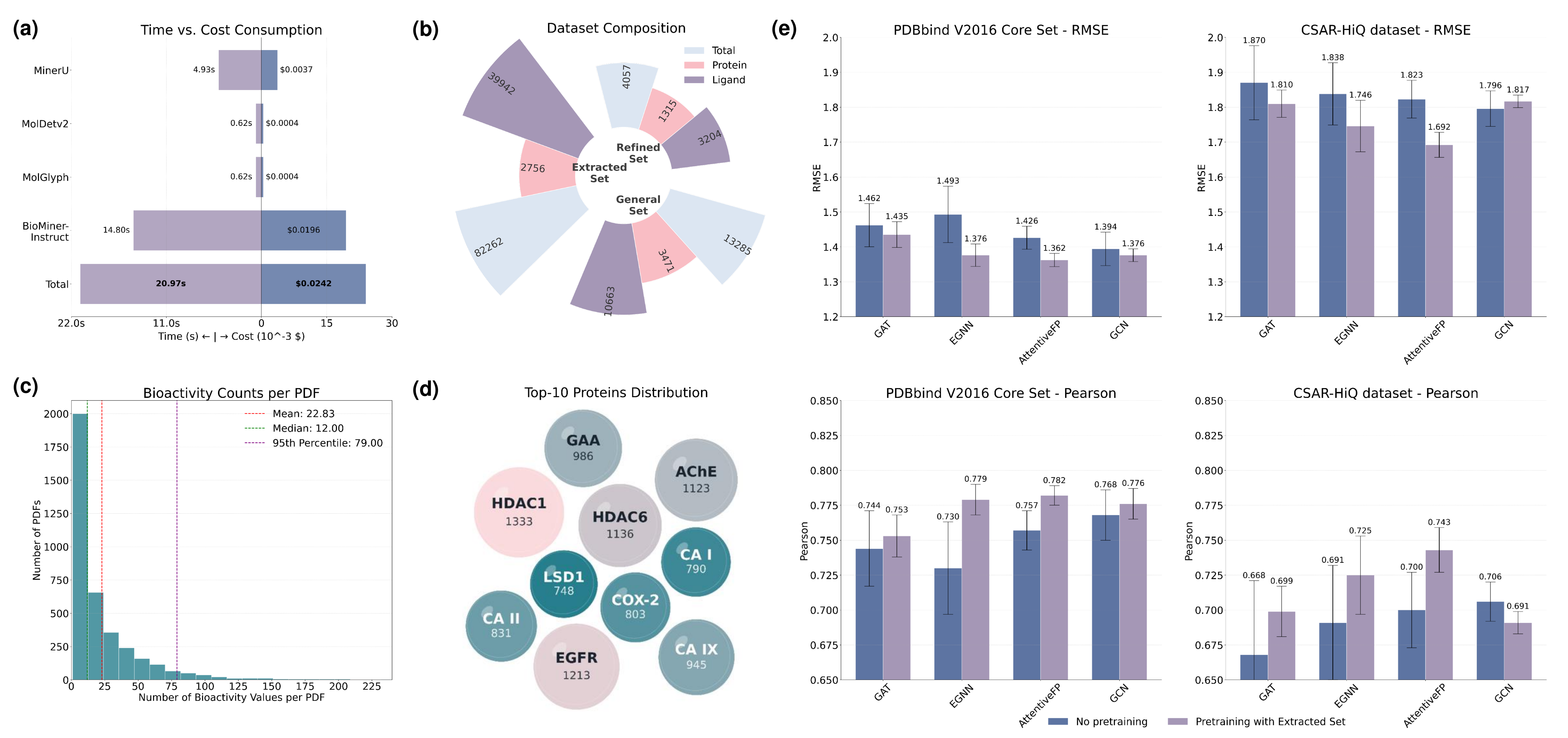}
\caption{
\yansr{Using \framework collecting large-scale bioactivity data for deep learning model training.
(a) Time and cost analysis of \framework.
(b) Statistical comparison between the manually curated PDBbind v2016 dataset and our extracted dataset.
(c) Number of extracted bioactivity data in each paper. \yansr{Papers without any bioactivity data are excluded.}
(d) The top-10 protein distribution within extracted bioactivity data.
(e) Performance comparison of models with and without pre-training on the \framework extracted set. Data are presented as mean ± SD from five independent runs ($n = 5$).
}
}
\label{fig:bioactivity_prediction}
\end{figure}

\vspace{-0.1cm}
\subsection*{Automated Collection of Large-scale Bioactivity Data for Model Training}
A key objective in developing \framework is to overcome the limitations of existing, manually curated bioactivity databases, and to unlock the wealth of previously inaccessible data hidden within scientific literature. To demonstrate \framework's capacity for large-scale data acquisition, we apply it to construct a bioactivity database. Deep learning-based bioactivity prediction models are further pre-trained on this extracted database to demonstrate their utility.

To construct a large-scale dataset, we target the European Journal of Medicinal Chemistry (EJMC), a high-impact journal known for its density of protein-ligand bioactivity data, making it an ideal source. 11,683 articles published in EJMC since 2010 are collected, excluding earlier articles due to potential low resolution. Employing \framework, we process the 11,683 papers within just \yan{three days (cost about 21 seconds and 0.024\$ per paper, as shown in Figure~\ref{fig:bioactivity_prediction}(\yansr{a}) and Table S4)}, a speed that would be practically impossible with manual curation.
\yan{From these papers, 226,076 bioactivity triplets are extracted. To support the training of bioactivity prediction models, we further enrich these data with protein structure information. Using the extracted protein name\yansr{s}, \framework searches external structure databases (including AlphaFoldDB\cite{varadi2024alphafold} and PDB\cite{burley2025updated}) for protein structure information.
82,262 data points are successfully enriched with protein structure information, providing a large-scale extracted database that significantly expands the available data compared to PDBbind refined and general sets (Figure~\ref{fig:bioactivity_prediction}(\yansr{b})).
After excluding paper\yansr{s without any} bioactivity data, analysis of the extracted dataset reveals a mean of 22.83 bioactivity values per paper (median = 12; 95th percentile = 79), highlighting the density of bioactivity data within individual publications (Figure~\ref{fig:bioactivity_prediction}(c)). Further analysis of the extracted dataset shows the top-10 protein distributions, indicating the most researched proteins and their potential application in bioactivity prediction (Figure~\ref{fig:bioactivity_prediction}(d)).}

Critically, we evaluate the impact of this automatically extracted database on the performance of downstream deep learning models. 
\yan{While the end-to-end extraction precision (\yan{0.32}) \yansr{implies} the presence of noise \yansr{in the extracted triplets}, deep learning models have shown remarkable robustness to massive label noise when trained on sufficiently large datasets~\cite{Rolnick2017DeepLI}. We hypothesize that the scale of our extracted data allows the model to "average out" the noise and capture generalized interaction features.}
We pre-train several graph neural network (GNN) architectures (GAT~\cite{Velickovic2017GraphAN}, EGNN~\cite{Satorras2021EnEG}, AttentiveFP~\cite{Xiong2020PushingTB}, and GCN~\cite{Kipf2016SemiSupervisedCW}) on the \framework-derived database and compare their performance \yansr{with} models trained solely on the PDBbind v2016 refined set \yansr{(details in Supplementary Note 3)}. Across two independent test sets (PDBbind v2016 core set and CSAR-HiQ set), \yansr{pre-trained models} consistently achieve performance improvements \yansr{in} both RMSE and Pearson correlation (Figure~\ref{fig:bioactivity_prediction}(e)). Average \yansr{RMSE reductions} of 3.9\% and 3.4\% are achieved, respectively.

\yan{To rigorously distinguish the value of the mined bioactivity signals from the benefit of simply scaling up structural data (unsupervised representation learning), we conduct additional comparisons against two baselines: (1) an unsupervised pre-training strategy that masks atoms/residues without using bioactivity labels, and (2) a negative control using shuffled bioactivity labels. 
As shown in Figure S8, the downstream performance follows a clear hierarchy: Shuffled << No pre-training < Unsupervised < Ours (\framework). Our method (3.9\% improvement) outperforms the unsupervised strategy (1.4\% improvement), confirming that the mined bioactivity values, despite inherent noise, provide unique \yansr{and} critical supervision signals regarding binding physics that cannot be learned from structural data alone.} 

These results underscore the value of automatically extracted bioactivity data for enhancing the accuracy and predictive power of computational models in drug discovery. \framework can serve as an efficient and cost-effective method for mining scientific literature, which is crucial for data-driven AI-powered drug design.

\begin{figure}[!htbp]
\centering
\includegraphics[width=1.0\linewidth]{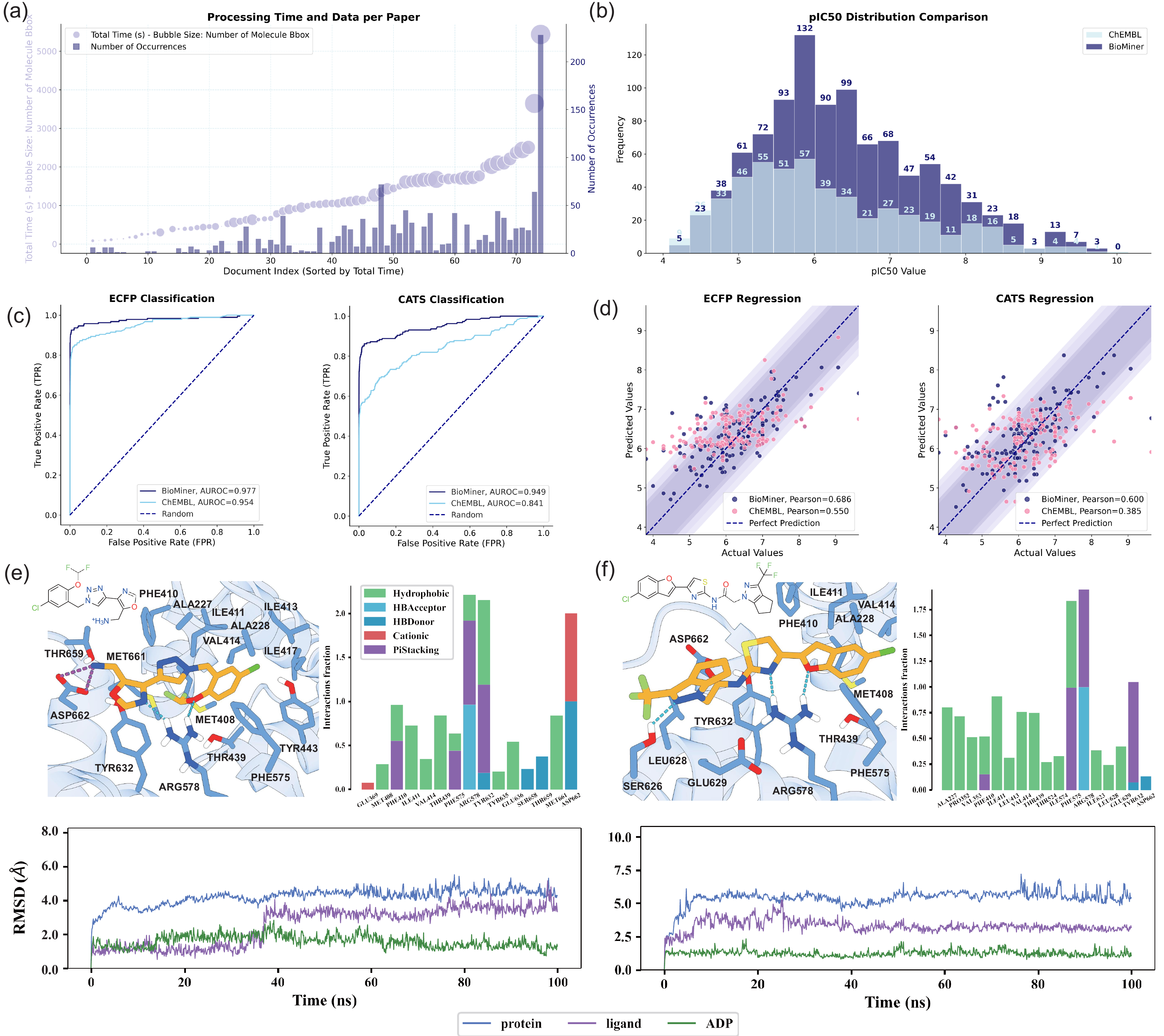}
\caption{
\yan{Using \framework for NLRP3 bioactivity data collection from 85 papers and inhibitor screening.
(a) Consuming time (about 18.4 minutes on average) and the number of collected data points for each paper. The final consuming time is highly related to the number of bioactivity data and chemical structures.
(b) Comparison of pIC50 distribution between \framework collected bioactivity data and ChEMBL data.
Classification (c) and regression (d) performance comparison between QSAR models trained on \framework data and ChEMBL data. The Glide-XP docked binding pose within the NP3-253-binding pocket (PDB ID: 9GU4), stacked bar plot of the interaction fraction during MD simulation, and RMSD curve of protein, ligand, and ADP for Z6739936901 (e) and Z5232931194 (f), respectively.}
}
\label{fig:nlpr3}
\end{figure}

\vspace{-0.1cm}
\subsection*{Human-in-the-loop Curation of NLRP3 Bioactivity Data and Enhanced Inhibitor Screening}
While fully automated extraction using \framework demonstrates high throughput, potential data errors will be introduced.
In a practical scenario, correcting potential errors and ensuring high data quality is crucial for sensitive downstream applications like QSAR model building and drug screening. 
To address this, we introduce a HITL workflow to enable high-quality and efficient data collection aided by \framework and validate its effectiveness in a practical task.

\subsubsection*{Bioactivity Data Collection HITL Workflow} 
In the HITL workflow, human experts extract bioactivity data by reviewing and validating the chemical structure and bioactivity measurement extraction results of \framework, which consists the final bioactivity triplet.
For chemical structures, as shown in Figure~S3, S5, and S6, a human expert verifies the first stage molecule detection, OCSR, and the second stage full structure coreference recognition, Markush enumeration, sequentially.
For bioactivity measurement, the expert directly confirms the correctness of the extracted protein-ligand coreference-bioactivity value tuples, the output of the bioactivity measurement extraction agent. 
Critically, human intervention is targeted only at these verification steps. Other components within \framework, such as 1D R-group processing and Markush scaffold-R-group zipping, are rule-based steps that do not require manual review.
Overall, this HITL workflow omits the \textit{de novo} identification and transcription of chemical structure and bioactivity, and experts primarily focus on verifying \framework's outputs, significantly accelerating curation compared to fully manual extraction.

\subsubsection*{NLRP3 Inflammasome Bioactivity Data Collection} 
To showcase this workflow's practical utility, we apply it to curate NLRP3~\cite{Swanson2019TheNI} inflammasome bioactivity data.
NLRP3 is a high-priority target for anti-inflammatory therapies, yet publicly available bioactivity data, such as in ChEMBL, remains relatively sparse. Our objective \yansr{is} to expand the size of high-quality NLRP3 bioactivity dataset and utilize this enriched data to develop more accurate models for inhibitor discovery.

We identify 85 relevant scientific publications reporting NLRP3 bioactivity data through literature searches and then collect 1,592 data from these publications with this HITL workflow.
The time required per paper \yansr{is correlated} with the number of chemical structures and bioactivity data points (Figure~\ref{fig:nlpr3}(a)). 
On average, it takes approximately 18.4 minutes for each paper and consumes 26 hours for all 85 papers.
This effort effectively doubles the amount of NLRP3 bioactivity data previously available in ChEMBL (Figure~\ref{fig:nlpr3}(b)), providing a substantially larger and chemically diverse dataset. The expert oversight inherent in the HITL process also allows for meticulous handling of ambiguities and edge cases, further ensuring data quality.

\subsubsection*{QSAR Models Training} 
To assess the impact of this data expansion, we train QSAR models for predicting NLRP3 inhibition using both the original ChEMBL NLRP3 dataset and our expanded, \framework-curated NLRP3 dataset, respectively. 
As shown in Table~S9 and Table~S10, various QSAR models with different algorithms (e.g., Random Forest~\cite{Breiman2001RandomF}, SVM~\cite{Hearst1998SupportVM}, etc.), different tasks (regression, classification), and different molecular representations (ECFP~\cite{Rogers2010ExtendedConnectivityF}, CATS~\cite{Reutlinger2013ChemicallyAT}) are evaluated comprehensively.
Models trained on the \framework-curated dataset consistently outperform those trained on the ChEMBL dataset across all settings. On average, considering the EF1\% metric across 28 distinct model configurations, the \framework-curated dataset yield\yansr{s} a 38.6\% performance improvement. The performance of the top-performing models under various settings \yansr{is visualized in} Figure~\ref{fig:nlpr3}(c, d). Notably, for classification using ECFP fingerprints, the AUROC improve\yansr{s} from 0.954 (ChEMBL) to 0.977 (\framework-curated). Similarly, for CATS-based regression, the Pearson correlation increase\yansr{s} substantially from 0.385 to 0.600. This marked enhancement in predictive performance significantly increases the reliability of bioactivity prediction for novel chemical entities, thereby improving prospects for effective inhibitor screening.

\subsubsection*{NLRP3 Inhibitor Screening} 
Leveraging the superior QSAR models trained on the \framework-curated data, we initiate a virtual screening against chemical libraries (i.e., ChemDiv and Enamine). 
Through our established rational screening pipeline (Figure~S11 and see Supplementary Note 5 for details), sixteen compounds are manually selected as potential hit candidates, exhibiting both high structural diversity and novelty (Table~S11).
Further MM/PBSA binding free energy calculation reveals six compounds exhibit comparable or better binding free energies than both MCC950~\cite{Coll2019MCC950DT} and NP3-562~\cite{Velcicky2024DiscoveryOP}, suggesting their potential as highly potent NLRP3 inhibitors~(Table~S11). 
Docking simulation demonstrates two promising candidates, Z6739936901 and Z5232931194, showing considerable binding mode between the NBD, HD1, WHD, and HD2 domains (Figure~\ref{fig:nlpr3}(e, f)). MD simulation of 100 ns is performed to investigate the binding stability of identified compounds. The RMSD of protein and ligand shows that Z6739936901 and Z5232931194 bind stably with the binding pocket in the last 60 ns. Z6739936901 maintains persistent hydrogen bonding, cation-$\pi$ and $\pi$-$\pi$ interactions with ARG578 and TYR632. Its solvent-exposed charged amino group forms strong salt bridges and hydrogen bonds with ASP662, anchoring the oxazole moiety rigidly. Meanwhile, the phenyl fragment remains tightly bound within a large hydrophobic subpocket. Z5232931194 frequently forms hydrogen bonds \yansr{and cation-$\pi$ with ARG578, and $\pi$-$\pi$ interactions} with PHE575 and TYR632. The bromine-substituted benzofuran moiety remains tightly bound within the hydrophobic subpocket. Notably, the ADP bound in the substrate pocket remains stable throughout MD simulations, with low RMSD values (< 1.5 \AA) confirming its rigid positioning with minimal conformational changes. This observation suggests that both Z6739936901 and Z5232931194 effectively stabilize the inactive conformation of the NBD, HD1, WHD, and HD2 domains. Collectively, our \framework-curated data-augmented QSAR models identifies several structurally novel hit candidates exhibiting high potential as potent NLRP3 inhibitors. These candidates represent promising chemotypes for future experimental validation, potentially unlocking new avenues in NLRP3-targeted drug discovery.

In summary, this study demonstrates the effectiveness of a \framework-powered HITL workflow for rapidly constructing large, high-quality, target-specific bioactivity datasets. The efficiency gain over manual curation, combined with the significant improvement in downstream QSAR model performance, underscores the potential of \framework to accelerate data-driven drug discovery efforts.

\begin{figure}[!t]
\centering
\includegraphics[width=0.94\linewidth]{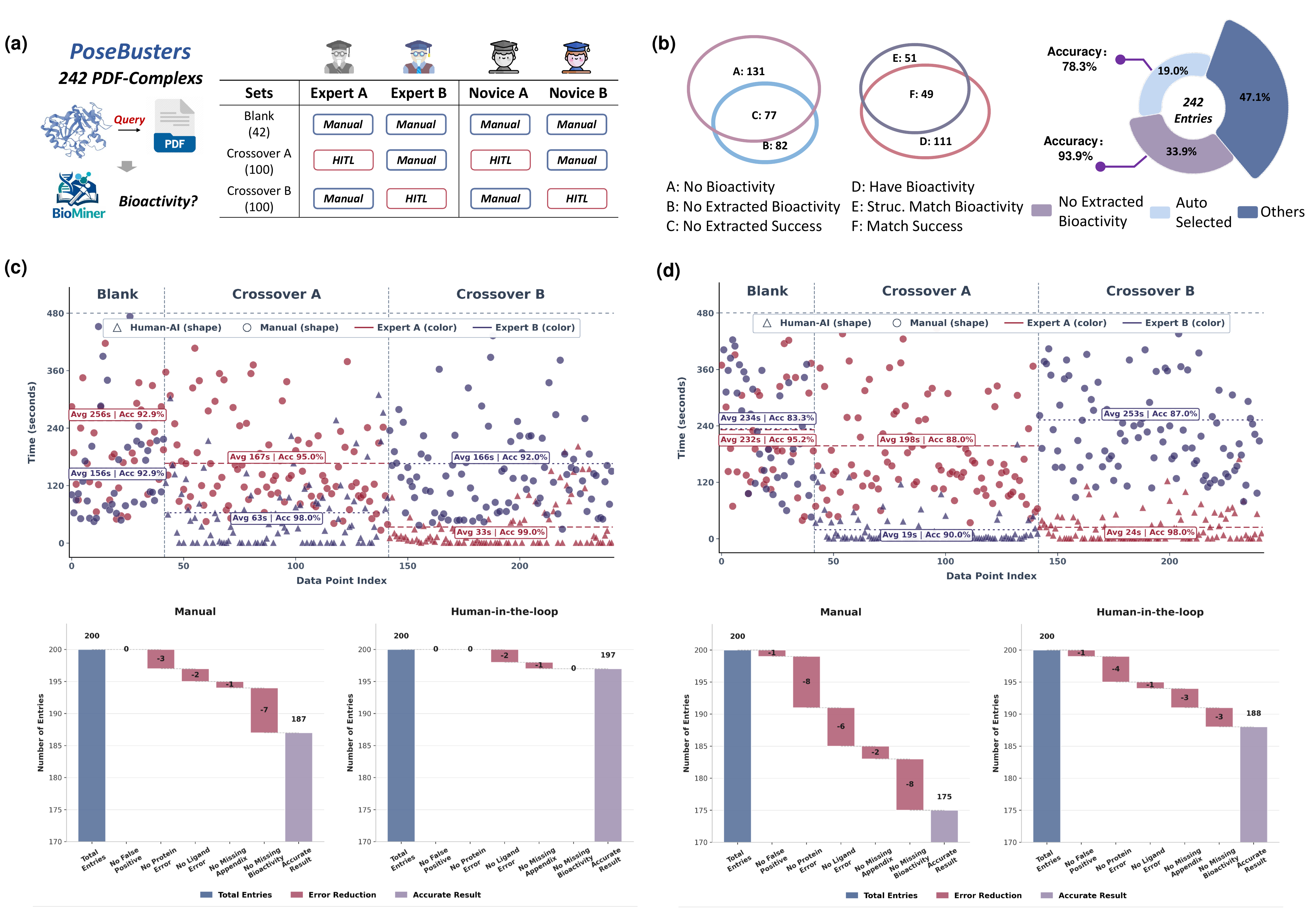}
\caption{
\yansr{Controlled evaluation of the \framework-assisted structure–bioactivity annotation on PoseBusters.
\textbf{(a)} Study design with one blank-baseline set (42 entries) and two crossover sets (100 entries each) from 242 test cases. 4 annotators, including 2 experts and 2 novices, are organized into matched pairs. 
\textbf{(b)} Fully automated annotation performance of \framework on the PoseBusters dataset.
\textbf{(c)} Per-entry annotation time and error decomposition of the expert group.
\textbf{(d)} Per-entry annotation time and error decomposition of the novice group. Full experimental details are provided in Supplementary Note 6.
}
}
\label{fig:posebuster}
\end{figure}

\vspace{-0.1cm}
\subsection*{Enhancing Structure-Bioactivity Annotation for PoseBusters} 
Besides bioactivity data extraction, annotating experimentally determined protein-ligand complex structures with their reported bioactivity measurements in the literature is also important in structure-based drug design field for establishing datasets like PDBbind.
In the experiments on the \benchmark, \framework has demonstrated its capability to perform this task. The reported bioactivity measurement of a given structure is often found among the top candidates of the extracted bioactivity data (top-10 recall = 0.598).
\yansr{Here}, we further demonstrate \framework's utility in a practical annotation scenario, proposing a HITL annotation workflow and presenting a detailed analysis for fully automated data annotation beyond Top-N Recall.

\subsubsection*{Structure-Bioactivity Annotation HITL Workflow} 
In the HITL annotation workflow, given a PDB complex structure and associated publication, \framework first extracts all potential protein-ligand \yansr{SMILES}-bioactivity triplets and ranks extracted data based on ligand similarity between extracted SMILES and PDB ligand SMILES.
Then, a human expert rapidly verifies the bioactivity data in the ranked list one by one.
Compared with purely human annotation, such a HITL streamlines the expert's task to validation rather than \textit{de novo} annotation.

\yansr{
\subsubsection*{Controlled evaluation on PoseBusters PDF-complexes}

We select the PoseBusters benchmark~\cite{Buttenschoen2023PoseBustersAD}, which contains 308 high-quality protein--ligand complex structures, as a realistic set for structure--bioactivity annotation.
After filtering for entries with accessible full-text PDF articles and usable inputs, 242 PDB--article pairs are retained for formal evaluation.

To explicitly control for inter-annotator variability, 4 annotators are recruited, including 2 expert annotators and 2 novice annotators.
The 242 cases are divided into one blank-baseline subset of 42 entries and two crossover subsets of 100 entries each (Figure~\ref{fig:posebuster}(a)).
On the blank-baseline subset, both annotators within each group perform fully manual annotation, providing a direct estimate of baseline variability under the same condition.
On the two crossover subsets, the annotators within each group alternate between manual annotation and the \framework-assisted HITL workflow, so that each annotator completes one subset under each condition.

The blank-baseline analysis confirms that annotator-specific variability is non-negligible (Figure~\ref{fig:posebuster}(c, d)), supporting the need for a controlled evaluation protocol.
Despite this baseline variability, the crossover evaluation consistently favors the HITL workflow.
In the expert group, HITL reduces annotation time from 167~s (166~s) to 63~s (33~s) per entry while improving accuracy from 95.0\% (92.0\%) to 98.0\% (99.0\%); in the novice group, it reduces annotation time from 198~s (253~s) to 19~s (24~s) while improving accuracy from 88.0\% (87.0\%) to 90.0\% (98.0\%) (Figure~\ref{fig:posebuster}(c, d)).
Aggregated over all crossover subsets, HITL improves final accuracy from 90.5\% to 96.25\% and reduces the average annotation time from 195.8~s to 35.0~s per entry, corresponding to a 5.59-fold speedup ($P<0.05$, univariate general linear model). 
Error decomposition further shows that the assisted workflow can reduce failures caused by missed bioactivity evidence, incorrect protein/ligand matching, and appendix-related omissions. 
}

Overall, these results show that \framework provides a practical and robust basis for structure--bioactivity annotation in a controlled HITL setting, improving both annotation efficiency and final annotation accuracy relative to fully manual curation.

\subsubsection*{Fully Automated Annotation} 
Based on the HITL annotation result, we further analyze the performance of the fully automated annotation beyond the Top-N Recall (Figure~\ref{fig:posebuster}(b)). 
For \yansr{82} of 242 complexes, \framework finds no bioactivity data in the publications. Among these complexes, \yansr{77} complexes are correct (consistent with manual verification confirming absence). 
For the remaining \yansr{160} complexes where bioactivity data are potentially present, \framework successfully extracts candidate bioactivity measurements. 
Within this group, \yansr{51} complexes have at least one extracted bioactivity measurement associated with a ligand structure perfectly matching the PDB ligand \yansr{, of which \yansr{49} are confirmed to be correct. }
For these matches, the data postprocessing agent (utilizing the Gemini MLLM) selects the most probable bioactivity value based on contextual consistency with the protein name and PDB structure title, \yansr{ identifying a candidate value in 46 cases.} Manual validation confirms that \yansr{36} of these 46 automatically selected annotations are correct, yielding an accuracy of \yansr{78.3\%} for this specific subset. Combining the \yansr{82} cases correctly identified as having no data with the 46 cases where a candidate value is proposed, \framework demonstrates reliable automated handling for \yansr{128} out of 242 structures (\yansr{52.9\%}). This suggests that nearly half of the annotation task for this dataset could potentially be automated with high confidence (Accuracy = \yansr{0.88}), significantly reducing manual effort, while the remaining cases benefit from the efficient HITL verification.

The ability to rapidly generate and verify high-fidelity structure-activity annotations for large structural datasets like PoseBusters is highly valuable for the development of structure-based drug design algorithms. 
By significantly accelerating the creation and curation of these critical datasets with enhanced accuracy, \framework directly facilitates progress in structure-guided drug discovery and the application of machine learning in structural biology.

\vspace{-0.2cm}
\section*{Discussion}
\yan{In this work, we introduced \framework, a multi-modal system for automated extraction of protein–ligand bioactivity data from scientific literature, built upon a principled integration of MLLMs, \yansr{DSMs}, and chemistry tools.
To support rigorous evaluation and foster systematic progress in this area, we further established \benchmark, the largest benchmark to date dedicated to protein–ligand bioactivity extraction, comprising 16,457 bioactivity entries and 8,735 chemical structures across six carefully designed evaluation tasks.

Evaluated on \benchmark, \framework demonstrated meaningful extraction capability under this highly challenging setting, achieving F1 scores of 0.32 for complete bioactivity triplets, 0.5\yansr{3} for chemical structure extraction, and 0.6\yansr{3} for bioactivity measurement extraction.
Beyond benchmark performance, the practical utility of \framework was substantiated through three real-world applications.
First, its scalability enabled the rapid construction of a large-scale bioactivity database (82,262 entries from 11,683 papers within three days), which yielded consistent improvements in downstream binding affinity prediction models, including a 3.9\% reduction in RMSE after pre-training.}
Second, HITL workflows highlighted the complementary strengths of automated extraction and expert validation: for the therapeutically important target NLRP3, HITL-assisted extraction doubled the volume of high-quality bioactivity data relative to ChEMBL within 26 hours, leading to substantially improved QSAR performance (38.6\% gain in EF1\%) and the identification of 16 novel hit candidates.
\yansr{Third, \framework proved effective for accelerating structure-centric annotation tasks, achieving a 5.59-fold speedup and higher accuracy (96.25\% versus 90.5\%) in labeling bioactivity-linked structures within the PoseBusters dataset.}

Despite these advances, the modest end-to-end F1 score for complete triplet extraction underscores the intrinsic difficulty of the task.
Protein–ligand bioactivity extraction requires precise cross-modal integration, reliable chemical structure reconstruction (particularly for Markush representations), and accurate association of entities and measurements scattered across heterogeneous document elements.
Our error analysis identifies several dominant bottlenecks, including bioactivity measurement extraction (\yan{32.68\%} of errors), \yansr{OCSR} (\yan{25.31\%}), and Markush structure enumeration (\yan{15.91\%}).
These findings indicate that performance limitations primarily stem from fundamental task complexity rather than superficial system design choices.
\yansr{Besides, regarding our architectural choice of modality separation via post-fusion, we acknowledge that our empirical comparison is limited by the fact that neither \framework nor the proxy model is optimized for an early fusion setting. Therefore, fully unified fusion architectures explicitly aligned for this multi-modal task remain a promising direction for future research to better exploit cross-modal synergies.}
\yansr{Finally, our evaluation design entails a trade-off with respect to the validation–test split. To maximize the robustness of our final evaluation, we restricted the validation set to 10\% of the data (50 papers). While sufficient for the limited prompt refinement performed in this study, this size leaves a relatively thin basis for extensive development decisions. Future work requiring more complex prompt engineering or extensive hyperparameter tuning would benefit from the curation of larger validation sets.
}

\yansr{Looking forward, the modular design of \framework suggests its potential for extension to related extraction tasks where exact chemical identity is required, such as extracting molecular ADMET properties from the literature. 
Although the current study does not yet examine cross-domain generalization, we believe \framework can serve as a transferable starting point, whose architecture may facilitate adaptation to related tasks with domain-specific adjustment.
}

\yan{In summary, \framework represents a step toward scalable, structure-aware automation of bioactivity data extraction, while \benchmark establishes a much-needed standard for systematic evaluation in this domain.
Together, they provide a foundation for unlocking large volumes of previously inaccessible bioactivity data embedded in literature, enabling more comprehensive data-driven modeling and accelerating future advances in drug discovery and automated scientific knowledge extraction.}

\clearpage

\clearpage
\section*{Methods}
\label{sec:method}
\subsection*{Preliminary}
\yan{For notation, a scientific article is represented as a multi-modal document $D$ (PDF), containing text, tables, and figures.
The goal is to extract a set of protein--ligand bioactivity records
$\mathcal{B}=\{(p_i, \ell_i, v_i)\}_{i=1}^{N}$,
where $p_i$ is a protein target mention, $\ell_i$ is a chemically valid ligand representation (SMILES), and $v_i$ is a quantitative bioactivity value with its associated assay context.

A key challenge is that bioactivity mining from literature \emph{structurally couples}:
(i) semantic reasoning over heterogeneous modalities (to interpret bioactivity measurements and entity relations) and
(ii) \emph{exact} symbolic construction of ligand structures (especially for Markush representations).
End-to-end generation is brittle.

\framework addresses this by decoupling:
\begin{equation}
\underbrace{\mathcal{M}(D)\rightarrow \{(p,c,v)\}}_{\text{bioactivity measurement: semantic reasoning}}
\quad \text{and} \quad
\underbrace{\mathcal{S}(D)\rightarrow \{(c,\ell)\}}_{\text{chemical structure: grounded + constrained construction}},
\end{equation}
and then integrating them via ligand coreference $c$:
\begin{equation}
\mathcal{B} = \mathcal{J}\big(\{(p,c,v)\}, \{(c,\ell)\}\big).
\end{equation}
Here $c$ denotes a ligand coreference string/identifier (e.g., compound name, label, or alias) used as the join key across modalities.
}

\subsection*{\textcolor{replyblue}{System overview}}
\yan{\framework is implemented as four specialized agents:
data preprocessing $A^{pre}$, chemical structure extraction $A^{str}$, bioactivity measurement extraction $A^{mea}$, and postprocessing/integration $A^{post}$.
At deployment, \framework runs inference-only: it orchestrates fixed-weight models and deterministic chemistry tools without document-specific training.

\begin{equation}
(D^{txt}, D^{vis}) = A^{pre}(D),\quad
\mathcal{S}=\{(c,\ell)\}=A^{str}(D^{vis}),\quad
\mathcal{M}=\{(p,c,v)\}=A^{mea}(D^{txt},D^{vis}),\quad
\mathcal{B}=A^{post}(\mathcal{S},\mathcal{M}).
\end{equation}

\paragraph{Backbone models and tools.}
\framework uses a locally deployable multi-modal language model \MLLM (fine-tuned from Qwen3-VL-32B) for cross-modal reasoning,
a domain-specific OCSR model \OCSR for chemical structure recognition in scientific figures,
and domain tools including RDKit for chemical graph operations and OPSIN for IUPAC-to-SMILES conversion.
(Implementation details and prompts are provided in Supporting Information.)

\subsection*{Agent 1: data preprocessing ($A^{pre}$)}
The preprocessing agent parses $D$ into aligned textual and visual representations while preserving layout and reading order.
We apply MinerU to obtain text and layout elements:
\begin{equation}
(D^{txt}, D^{lay}) = \texttt{MinerU}(D),
\end{equation}
where $D^{txt}$ contains extracted text segments and $D^{lay}$ stores page-level layout elements with their bounding boxes and categories (e.g., paragraph, table, figure, caption).

For visual grounding, each page is rasterized into an image and we additionally extract figure/table crops when possible:
\begin{equation}
D^{vis} = \big(D^{vis}_{page}, D^{vis}_{seg}\big).
\end{equation}
$D^{vis}_{page}$ contains full-page images, which are robust to imperfect caption detection and support molecule detection at page scale.
$D^{vis}_{seg}$ contains segmented figure/table regions (with captions when detected) to reduce surrounding noise for downstream visual reasoning.

\subsection*{Agent 2: chemical structure extraction ($A^{str}$)}
$A^{str}$ produces a set of ligand coreferences and their chemically valid SMILES:
\begin{equation}
\mathcal{S}=\{(c_k,\ell_k)\}_{k=1}^{K} = A^{str}(D^{vis}).
\end{equation}

Chemical structures in literature appear as:
(i) explicit full structures, and
(ii) Markush structures that define families of compounds via variable substituents (R-groups).
Our scope focuses on Markush instances where the scaffold and attachment points are representable in SMILES; other Markush types (e.g., variable positions/repeats not representable in SMILES) are handled as out-of-scope and reported separately.

CSG-VSR resolves chemical structures through a three-stage grounded pipeline:
\begin{equation}
\text{(detect + recognize)} \rightarrow
\text{(grounded relation inference)} \rightarrow
\text{(tool-constrained symbolic construction)}.
\end{equation}

\paragraph{Stage I: molecule detection and OCSR.}
We detect 2D chemical depictions in $D^{vis}_{page}$ and $D^{vis}_{seg}$ and recognize each depiction into a preliminary SMILES string.
Let $\{(x_m,\tilde{\ell}_m)\}_{m=1}^{M}$ denote detected bounding boxes $x_m$ with OCSR outputs $\tilde{\ell}_m$.
Depictions with attachment points are tagged as Markush components; otherwise they are treated as explicit full structures.

\paragraph{Stage II: grounded relation inference (coreference + Markush enumeration).}
To reliably refer to visual structures, we create an augmented image by overlaying bounding boxes and unique indices on each detected depiction.
Given an augmented image $I^{aug}$, \MLLM infers:
\begin{itemize}[leftmargin=*]
\item \textbf{Coreference for explicit structures:} mapping from visual indices to ligand identifiers $c$ (e.g., ``compound 12'', ``AZD1234'', ``R1'').
\item \textbf{Markush enumeration:} for each Markush scaffold, identify the scaffold index and enumerate substituent sets (R-groups), where substituents can be (a) visual indices of depicted fragments or (b) textual 1D mentions (IUPAC, abbreviations, formulas).
\end{itemize}

\paragraph{Stage III: tool-constrained symbolic construction.}
We convert R-group substituents into SMILES:
IUPAC names via OPSIN, and abbreviations/formulas via a curated mapping table (assisted by \MLLM during construction; the finalized table is fixed for deployment).
Finally, RDKit composes scaffold and substituent SMILES by attachment-point ``zipping'' to generate chemically valid full structures.
This stage enforces chemical validity deterministically and isolates the MLLM from error-prone string construction.

The final output $\mathcal{S}$ merges:
(i) explicit structures $(c,\ell)$ and
(ii) Markush-enumerated structures $(c,\ell)$ obtained by RDKit construction.

\subsection*{Agent 3: bioactivity measurement extraction ($A^{mea}$)}
$A^{mea}$ extracts bioactivity measurements as semantic tuples $(p,c,v)$ from both text and visual modalities:
\begin{equation}
\mathcal{M}=\{(p_t,c_t,v_t)\}_{t=1}^{T}=A^{mea}(D^{txt},D^{vis}).
\end{equation}
We run \MLLM over (a) $D^{txt}$ and (b) page-level images $D^{vis}_{page}$ to capture measurements reported in paragraphs, tables, and figure panels.
Outputs are normalized into a unified schema, including value types (e.g., IC$_{50}$, K$_i$), numeric values, and units when available.
Text- and vision-derived tuples are then merged with deduplication based on entity/value consistency rules (Supporting Information).

\subsection*{Agent 4: postprocessing and integration ($A^{post}$)}
The postprocessing agent joins $\mathcal{M}$ and $\mathcal{S}$ via ligand coreference to produce final triplets:
\begin{equation}
\mathcal{B} = A^{post}(\mathcal{S},\mathcal{M}) = \{(p,\ell,v)\}.
\end{equation}

\paragraph{Optional enrichment.}
For downstream usability, $A^{post}$ optionally enriches records with external identifiers and structures by querying public databases (e.g., UniProtKB, PDB, AlphaFoldDB), and can link extracted bioactivities to reported PDB complexes by SMILES matching and target consistency checks.

\paragraph{Summary.}
Overall, \framework operationalizes a structure-grounded paradigm: \MLLM performs cross-modal \emph{relation inference} (what refers to what, which R-groups belong where, which measurements belong to which targets),
while \yansr{DSMs} provide perception (OCSR/detection) and domain tools guarantee \emph{exact} chemical symbolic construction and validity.
}

\yan{
\subsection*{Model training for \framework}
\label{sec:model_training}
\framework is an inference-only agentic system at deployment time; however, we develop two open-weight models to enable fully reproducible local execution: (i) an OCSR model \OCSR for robust recognition of complex chemical depictions in scientific figures, and (ii) a bioactivity-oriented multimodal instruction model \MLLM for grounded cross-modal reasoning (coreference, Markush enumeration, and image-based bioactivity parsing). Importantly, the training of these models is \emph{document-independent} and strictly isolated from \benchmark{} to avoid data leakage.

\subsubsection*{Training OCSR \OCSR}
\paragraph{Architecture.}
\OCSR follows an encoder--decoder OCSR design instantiated from MolScribe, using a Swin-Transformer visual encoder and an Transformer decoder with 12 layers, 16 attention heads, and 512 hidden dimension.

\paragraph{Tokenizer and data format.}
We implement the tokenizer and sequence format following the MolParser paper specification.
MolParser does not provide an open-source implementation; therefore, all tokenization, rendering, and training pipelines are independently implemented in this work, while remaining compatible with the released MolParser pretraining data.

\paragraph{Training Data.}
We use three complementary data sources:
(1) \textbf{MolParser pretraining set} (8M images), which contains $\sim$2M synthesized Markush-like samples and provides broad coverage of depiction styles;
(2) \textbf{MolParser real-world set} (91k images) as a curated evaluation-oriented distribution of real depictions;
(3) \textbf{Literature-specific OCSR set} (170k images) constructed in this work to mitigate domain shift from patents to scientific articles.
The MolParser corpus is predominantly patent-derived, which differs from the visual style and composition of biomedical literature (e.g., tighter layouts, multi-panel figures, smaller line widths, and heavier co-occurrence with biological annotations).
To bridge this gap, we collect chemical depiction panels from (i) PDBbind-cited publications prior to 2019 and (ii) high-bioactivity-density medicinal chemistry journals (JMC, JNP, BMC, BMCL).
To prevent evaluation leakage, we ensure that journals/papers included in \benchmark{} are excluded from this OCSR training pool; additionally, to reduce redundancy, we sample 3--5 figure panels per paper and cap the number of samples per article source when necessary.
Among the three data sources, the literature-specific set is weakly labeled by MolParser (closed API) to obtain paired (image, SMILES) supervision. We treat these labels as \emph{pseudo-ground-truth} and rely on large-scale diversity + curriculum learning (below) to reduce sensitivity to occasional labeling noise.

\paragraph{Curriculum learning and augmentation.}
\OCSR is trained with a curriculum strategy: we first pretrain on the large-scale MolParser pretraining set, and then progressively introduce stronger augmentations and harder samples.
After convergence, we perform joint training on a mixed corpus consisting of: the MolParser real-world 91k set, a balanced subset of $\sim$200k synthesized Markush samples, and the 170k literature-specific set.
This final mixture emphasizes the scientific-figure distribution while retaining sufficient Markush coverage.

\paragraph{Optimization and hyperparameters.}
The model is trained end-to-end using teacher-forced cross-entropy loss with label smoothing of 0.1.
We use the AdamW optimizer with separate learning rates of $4\times10^{-4}$ for both encoder and decoder.
A linear warm-up schedule is applied for the first 2\% of training steps, followed by cosine decay.
Training is conducted for 10 epochs over the combined corpus.
To improve stability on noisy pseudo-labeled data, we filter invalid structure graphs during training and apply SMILES syntax constraints during decoding.
It takes 7 days to train \OCSR on 4 A800 80G GPUs.

\subsubsection*{Training MLLM \MLLM}
\paragraph{Base model.}
We initialize \MLLM from Qwen3-VL-32B-Instruct and perform multi-task supervised instruction tuning (SFT) on a curated multi-modal dataset covering three tasks used by \framework:
(1) \textbf{Markush enumeration} (scaffold identification and R-group set enumeration),
(2) \textbf{Full-structure coreference recognition} (mapping depicted structures/fragments to textual identifiers),
(3) \textbf{Image bioactivity extraction} (interpreting dense quantitative bioactivity values).

\paragraph{Instruct-tunning Data.}
All candidate papers are drawn from PDBbind-cited publications prior to 2019 and the same four high-bioactivity journals (JMC, JNP, BMC, BMCL), with explicit exclusion of the 500 \benchmark{} papers.
This isolation is enforced at the paper level (no shared PDF sources).
We first run a proprietary MLLM (Gemini-2.0-flash) to generate candidate input--output pairs for all three tasks.
Then, we filter and mine difficult examples to avoid over-representing trivial instances. For Markush enumeration, we discard invalid outputs and cases enumerating fewer than 8 structures. For image bioactivity extraction we prioritize dense figures. And for coreference recognition, since \framework adopts a divide-and-conquer strategy with a controlled number of molecules per image (typically $\le$ 4), we select data randomly.
Followed by human expert checking to ensure data quality, the final multi-task SFT set contains approximately 754 Markush enumeration instances, 2,215 coreference instances, and 4,375 image bioactivity instances.

\paragraph{Fine-tuning strategy.}
\framework is obtained by supervised instruction tuning of Qwen3-VL-32B-Instruct using parameter-efficient low-rank adaptation (LoRA)~\cite{Hu2021LoRALA}.
We adopt 4-bit quantization during training (QLoRA)~\cite{Dettmers2023QLoRAEF} to reduce GPU memory usage while preserving model capacity.
The instruction dataset jointly covers three tasks: Markush enumeration, full-structure coreference recognition, and image-based bioactivity extraction. These three tasks are mixed uniformly during training without task-specific weighting.

\paragraph{Optimization and hyperparameters.}
Training is performed using the AdamW optimizer with a peak learning rate of $5\times10^{-5}$.
A cosine learning-rate scheduler with a warm-up ratio of 0.1 is applied.
It takes one day to train \MLLM on 4 A800 80G GPUs.

}

\section*{Acknowledgements}
We sincerely thank Xi Fang from DP Technology for their support of MolParser.
We also extend our gratitude to Ruikang Li and Qingchuan Li from USTC, as well as Qian Yan, Qian Xie, Yiwen Zhang, Tongtong Yan, Huangdong Liang, and Li Wang from iFLYTEK, for their invaluable support in the construction of the benchmark \benchmark. 

\section*{Author contributions statement}
Conceptualization: J.X.Y., X.K.L., J.T.Z., Z.X.Z., and Q.L.;
Data Curation: J.X.Y., K.Z., Y.H.Y., and B.Y.Z.; 
Formal Analysis: J.X.Y., Y.H.Y., K.Z., Z.X.Z., J.T.Z., and X.K.L.;
Investigation: J.X.Y., and J.T.Z.;
Methodology: J.X.Y., X.K.L., Z.X.Z., and J.T.Z.;
Software: J.X.Y., Y.H.Y., J.T.Z., and J.C.X.;
Validation: J.X.Y., J.T.Z., K.Z., Z.X.Z., and Q.L.;
Visualization: J.X.Y., and J.T.Z.;
Writing – Original Draft: J.X.Y., and J.T.Z.;
Writing – Review \& Editing: J.X.Y., J.T.Z., Z.X.Z., K.Y.G., and Q.L.;
Supervision: Z.X.Z., K.Z., and Q.L.;
Project administration: K.Z., and Q.L.;
Resources: K.Z., K.Y.G., and Q.L.;
Funding Acquisition: Q.L.;
All authors reviewed the manuscript.

\section*{Data availability}
The proposed benchmark \benchmark is available in \url{https://github.com/jiaxianyan/BioMiner}. 
The PDBbind dataset is available in \url{http://www.pdbbind.org.cn/}. 
The ChEMBL dataset is available in \url{https://www.ebi.ac.uk/chembl/}. 
The molecule library ChemDiv is available in \url{https://www.chemdiv.com/}. 
The molecule library Enamine is available in \url{https://enamine.net/}. 
The PoseBuster dataset is available in \url{https://github.com/maabuu/posebusters}. \yansr{All source data, including extracted bioactivity dataset from EJMC papers, extracted NLRP3 bioactivity data and annotated structure-bioactivity pairs for PoseBusters set are provided in \url{https://github.com/jiaxianyan/BioMiner}}.

\section*{Code availability}
All codes \yansr{and model weights} of proposed \framework are released \yansr{and publicly available at} \url{https://github.com/jiaxianyan/BioMiner}. \yansr{The code was released under the MIT license.}

\clearpage
\bibliography{sample}

\end{document}